\documentclass[runningheads]{llncs}

\usepackage{xcolor}
\usepackage{graphicx}
\usepackage{mathtools}
\usepackage{multirow}
\usepackage{algorithm}
\usepackage{algorithmic}
\usepackage{adjustbox}
\usepackage{tikz}
\usepackage{pgfplots}
\usepackage{ctable}
\begin{document}
\title{On-Device Soft Sensors: Real-Time Fluid Flow Estimation from Level Sensor Data}
\titlerunning{On-Device Soft Sensors for Fluid Flow Estimation}
\author{Tianheng Ling\inst{1}\orcidID{0000-0003-4603-8576}
\and
Chao Qian\inst{1}\orcidID{0000-0003-1706-2008} 
\and
Gregor Schiele\inst{1} \orcidID{0000-0003-4266-4828}}
\authorrunning{T. Ling et al.}
\institute{Intelligent Embedded Systems Lab, University of Duisburg-Essen \\ 47057 Duisburg, Germany \\
\email{\{tianheng.ling, chao.qian, gregor.schiele\}@uni-due.de} 
}
\maketitle           
\begin{abstract}

Soft sensors are crucial in bridging autonomous systems' physical and digital realms, enhancing sensor fusion and perception. Instead of deploying soft sensors on the Cloud, this study shift towards employing on-device soft sensors, promising heightened efficiency and bolstering data security. Our approach substantially improves energy efficiency by deploying Artificial Intelligence (AI) directly on devices within a wireless sensor network. Furthermore, the synergistic integration of the Microcontroller Unit and Field-Programmable Gate Array (FPGA) leverages the rapid AI inference capabilities of the latter. Empirical evidence from our real-world use case demonstrates that FPGA-based soft sensors achieve inference times ranging remarkably from 1.04 to 12.04 $\mu$s. These compelling results highlight the considerable potential of our innovative approach for executing real-time inference tasks efficiently, thereby presenting a feasible alternative that effectively addresses the latency challenges intrinsic to Cloud-based deployments.

\keywords{On-Device AI\and 
Soft Sensors \and
Real-Time Inference \and
FPGA-Based Inference}
\end{abstract}

\section{Introduction}

Soft sensors, envisioned as virtual constructs, utilize algorithms and computational models to infer impalpable values by processing data derived from physical sensors~\cite{becker2010softsensorsysteme}. They have gained traction in ubiquitous computing, offering invaluable assistance where obtaining direct measurements is either impractical or economically burdensome, enriching data-driven insights in various systems~\cite{graziani2020deep}. Integrating Deep Learning (DL) techniques into soft sensors has spurred a movement toward Cloud-based deployments~\cite{graziani2020deep}. Since these soft sensors function as services on the Cloud, they inherit undeniable advantages: scalable computational resources, vast storage for training data, enhanced data analytic capabilities for model refinement, and flexible access to data and insights~\cite{sun2021survey}. 

However, Cloud service effectiveness is closely linked to the stability and reliability of network conditions. Even slight instability in network connections can significantly delay the transmission and processing of sensor data, adversely affecting the real-time applications that rely on swift and accurate sensor data for immediate decision-making tasks. Furthermore, the low transmission rates of many wireless sensor networks present substantial challenges for the continuous, real-time transmission of extensive sensor data to the Cloud. For example, LoRaWAN achieves transmission rates of 0.3 kbps to 50 kbps~\cite{phung2018analysis}.

In response to these pressing challenges, this paper promotes a shift towards implementing on-device soft sensors. We introduce a novel approach that seamlessly integrates low-power Microcontroller Units (MCUs) with embedded Field-Programmable Gate Arrays (FPGAs). This integration allows for the execution of AI functionalities directly on devices, thereby reducing latency, enhancing reliability, and bolstering real-time system performance. These improvements collectively offer a robust and efficient alternative to the prevalent issues associated with Cloud-based approaches. To substantiate the viability and efficacy of this approach, this study delves into a practical application: estimating fluid flow using level sensor data. Furthermore, we evaluate model performance, energy efficiency, and timing on both MCU and FPGA platforms, highlighting the tangible benefits of our innovative strategy in real-world scenarios.

The remainder of this paper is structured as follows: Section \ref{sec:related_work} reviews the literature relevant to our study. Section \ref{sec:system_architecture} outlines the proposed system architecture, while Section \ref{sec:case_study} presents a case study demonstrating the effectiveness of our approach. Finally, Section \ref{sec:conclusion} concludes the paper and discusses future research.

\section{Related Work}
\label{sec:related_work}

The burgeoning soft sensors have integrated into various domains~\cite{sharma2022recent}. Despite their invaluable capabilities, deploying these sensors in Cloud-based environments invites challenges, predominantly network latency and data security, which often diminish their benefits, necessitating effective alternatives.

Recent research signals a shift towards on-device soft sensors. Aguasvivas Manzano et al.~\cite{aguasvivas2022toward} exemplified this by deploying Neural Network models on ARM Cortex-M4 MCUs, realizing soft sensors with 7ms inference time and 45.4 mW power consumption. Flores et al.~\cite{flores2022tinyml} applied Tiny ML-based soft sensors in vehicles, implementing a Multi-layer Perceptron (MLP) model on a 32-bit MCU (ESP32) for efficient CO2 emissions estimation, showcasing a decreased model inference time to 37 \(\mu\)s by tweaking the model size. Balaji et al.~\cite{balaji2023ai} introduced a novel approach, integrating reconfigurable hardware accelerators for fast on-device AI inference in a wearable interface, with initial prototypes using FPGAs for emulation, marking significant improvements in speed and efficiency. 

Drawing inspiration from these works, we introduce an innovative approach for implementing on-device soft sensors by integrating MCUs with embedded FPGAs. To validate this, we focused on estimating fluid flow using level sensor data, a critical task in our research project, as directly using flow sensors in underground sewer systems introduces challenges due to their high cost, maintenance demands, and vulnerability to harsh conditions~\cite{tomperi2022estimation}. While traditional physical and statistical models have been extensively utilized, the complexity of real-world phenomena, amplified by factors like human activities, poses challenges for traditional linear models to capture non-linear and intricate patterns. There is a discernible gap in the literature concerning DL-based research, with a need for dedicated public data sets exacerbating the issue. Notably, even fewer studies explore on-device AI inference to address this demand.

We commenced our research with the data set released by Noori et al.~\cite{noori2020non}. Although limited in size, it offers a starting point for our investigation. Using an MLP model with a non-contact radar sensor to estimate non-Newtonian fluid flow in open Venturi channels, their work provides a foundational baseline. Our study further explores the potential of on-device soft sensors for real-time data processing tasks, responding to the challenges presented by Cloud-based IoT deployments, and aims to contribute to the ongoing discourse in the field.

\vspace{-5pt}
\section{System Architecture}
\label{sec:system_architecture}
This section presents the architecture of our on-device soft sensor system. We begin with defining and elucidating soft sensors, then delve into the process of localizing these sensors on target hardware. This examination will highlight our proposed system architecture's distinctive features and advantages.

\subsection{Architecture Overview}
\begin{figure}[!htb]
\vspace{-10pt}
    \centering
    \includegraphics[width=0.70\textwidth]{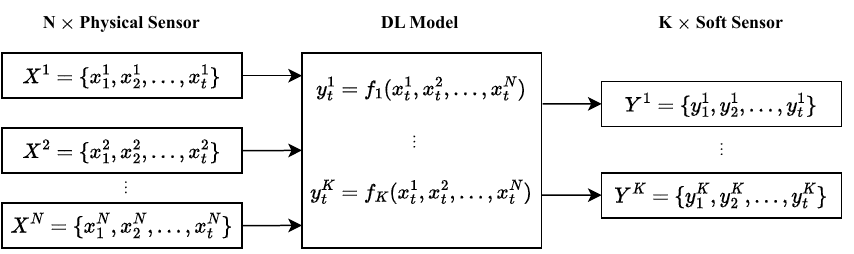}
    \caption{General System Architecture of Soft Sensors}
    \label{fig:LocalizedSoftSensor}
    \vspace{-10pt}
\end{figure}
Figure \ref{fig:LocalizedSoftSensor} presents our proposed general system architecture, designed to generate \( K \) independent soft sensors from data collected by \( N \) sensors. In this architecture, \(X^1\) to \(X^N\) represent discrete time series obtained by sampling \( N \) sensors at consistent intervals of \( T \). We introduce \( K \) independent fusion functions, denoted as \( f_1 \) through \( f_K \), each designed to integrate the \( N \) inputs at any given time \( t \). These functions yield new features serving as the output for specific soft sensors at each time point. Maintaining the initial sampling frequency, the architecture illustrated in Figure \ref{fig:LocalizedSoftSensor} effectively transforms an N-dimensional time series into a K-dimensional one. Crucially, when \(K\) is less than \(N\), this results in a reduction of the data volume required for wireless transmission, diminishing it to a \(K/N\) fraction of the original volume. These fusion functions must be completed within period T to meet the real-time requirement.

In our approach, these fusion functions are actualized within the proposed system by implementing DL models. The training phase of these models occurs on the Cloud during the initial development stage of our soft sensor system. After attaining satisfactory training outcomes, the models are deployed onto IoT devices, thereby operating as efficient on-device soft sensors.

\subsection{Elastic Node AI Hardware Platform}

\begin{figure}[!htb]
  \centering
  \begin{minipage}{0.3\columnwidth}
    \centering
    \rotatebox{90}{\includegraphics[width=60pt]{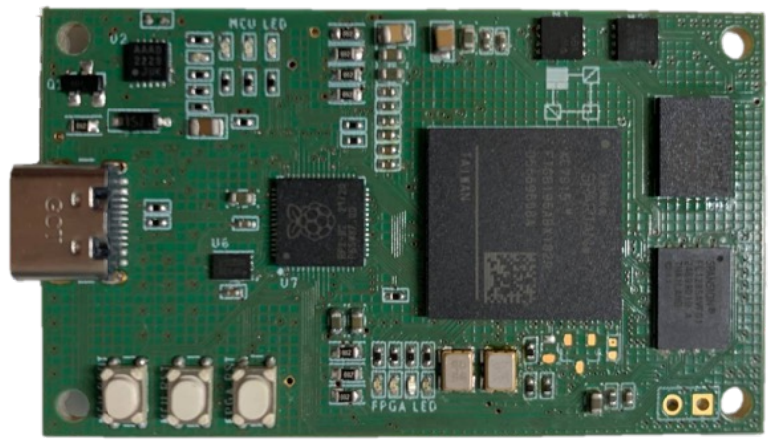}}
    \caption{Elastic Node V5}
    \label{fig:SensorOnDevice2}
  \end{minipage}
  \hfill
  \begin{minipage}{0.65\columnwidth}
    \centering
    \includegraphics[height=60pt]{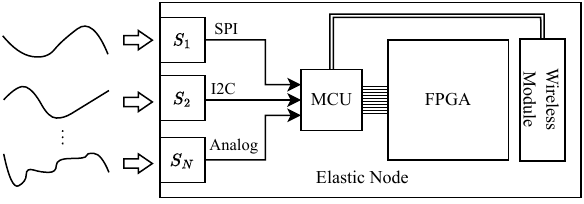}
    \caption{Elastic Node V5 with Soft Sensors}
    \label{fig:SensorOnDevice1}
  \end{minipage}
  \vspace{-10pt}
\end{figure}

Figure \ref{fig:SensorOnDevice2} illustrates the fifth version of the \textit{Elastic Node}, a hardware platform meticulously designed for on-device AI development~\cite{qian2023elasticai}. This platform primarily employs the Cortex-M0 MCU (RP2040), efficiently controlling integrated resources like sensors, a wireless module, and the Spartan-7 S15 FPGA. The FPGA plays a pivotal role, serving as configurable hardware for DL accelerators to significantly enhance the speed and energy efficiency of DL inference tasks. For detailed hardware insights, refer to Figure \ref{fig:SensorOnDevice1}, which provides a schematic of the soft sensor's hardware implementation. The sensors, labeled \(S_1\) to \(S_N\), interface with the MCU via a digital data bus (e.g., I2C or SPI) or analog channels connected to the MCU's internal ADC.

Regarding operational dynamics, the FPGA remains inactive until the MCU activates it upon accumulating sufficient raw data. Depending on the computational needs, the MCU either directly proceeds with model inference or delegates this task to the accelerator, necessitating the FPGA's activation and configuration. After processing, the soft sensor's output can be wirelessly transmitted to the Cloud. Furthermore, due to its reconfiguration feature, the FPGA can seamlessly switch between various DL models.

\subsection{Model Deployment on Elastic Node}

\begin{figure}[!htb]
    \centering
    \includegraphics[height=48pt]{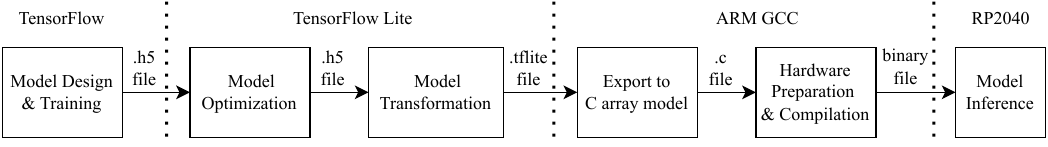}
    \caption{Workflow of DL Deployment on MCU}
    \label{fig:workflow1}
    \vspace{-10pt}
\end{figure}

Leveraging Elastic Node V5's heterogeneous architecture, which includes MCUs and FPGAs, our soft sensors can execute on either platform, providing deployment flexibility. Figure \ref{fig:workflow1} illustrates the simplified training and deployment process of DL models onto MCUs using TensorFlow (TF) Lite, facilitating a seamless transition from development to implementation.

 We initiate the process by designing and training a DL model for a specific application and performance criterion. For instance, the Mean Square Error (MSE) between the model output and test label must be within a set threshold. Using TF Lite's APIs, we can optimize the model using techniques like quantization and pruning for MCU deployment. Due to the absence of file system support on MCU, the model should be converted from the \textit{.tflite} format to a C array. The C array is then incorporated into a project with runtime programs and dependencies, facilitating the generation of an executable binary compatible with the MCU. This process culminates in the binary's firmware to the MCU for on-site computations.

\begin{figure}[!htb]
\vspace{-10pt}
    \centering
    \includegraphics[height=55pt]{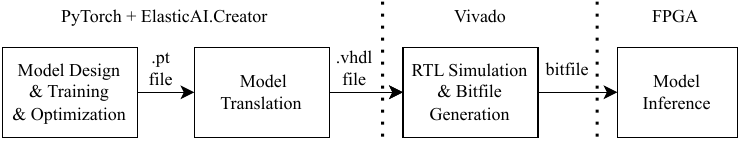}
    \caption{Workflow of DL Deployment on FPGA}
    \label{fig:workflow2}
    \vspace{-10pt}
\end{figure}

Crafting efficient FPGA hardware accelerators is intricate. To simplify this challenge, we employed the \textit{ElasticAI.Creator} toolchain~\cite{qian2022elasticai}, tailored for the PyTorch (PT) framework, as illustrated in Figure \ref{fig:workflow2}. This tool facilitates the intuitive design of models using its supported layer components. The toolchain manages the model's training and optimization to ensure it is suited for the embedded FPGA specifications. As a result, it outputs VHDL files representing the hardware accelerator at the Register-Transfer Level (RTL). These RTL designs are subsequently synthesized by the Vivado design suite into a bitfile. With the capacity to store multiple bitfiles on one flash, the MCU can direct the FPGA to switch between them as needed, thereby emulating different soft sensors. In this configuration, the MCU predominantly handles physical sensor data transmission to the FPGA and triggers the inference process, while the FPGA undertakes the computational-heavy tasks, signaling the MCU once finished.

\section{Case study: Fluid Flow Estimation}
\label{sec:case_study}
This section delves into a case study centered on real-time fluid flow estimation via level sensor data. We begin by describing the data set used, then outline our experimental setup, and conclude by presenting and evaluating the results garnered from our approach.

\subsection{Data Set and Experiments Settings}
Noori et al.~\cite{noori2020non} simulated drilling conditions using an advanced flow loop apparatus comprising a mud tank, blender, centrifugal pump, and an open Venturi channel. Three level sensors were utilized: an ultrasonic level meter, a VEGAPULS 64 radar, and a KROHNE Optiwave radar. The data from these sensors served as input data, while a precise Coriolis mass flow meter provided the target data. Data was collected at a 10kHz sampling rate. We used 1,800 samples from their study, allocating 80\% for training and 20\% for testing.

We employed a simple MLP model analogous to the one used in the study by Noori et al.~\cite{noori2020non}, comprising one hidden layer and an output layer with a single neuron. Experiments were conducted with hidden layers of 10 (as initially used), 30, 60, and 120 neurons to investigate the impact of model size on performance, to deploy on-device soft sensors on both MCU and FPGA. 

For MCU-based soft sensors, training was conducted under the TF framework, utilizing the \textit{Adam} optimizer with parameters \( \beta_1=0.9 \), \( \beta_2=0.98 \), and \( \epsilon=10^{-9} \). An initial learning rate of 0.001, step size of 3, and gamma of 0.5 were set for learning rate scheduling. We used the MSE loss function and also employed MSE as the evaluation metric. After executing 100 experiments across 500 epochs with early stopping, the optimal full-precision (\( \text{TF}_{FP32} \)) model  was selected. This model was subsequently quantized to 8-bit integer formats (\( \text{TF}_{quant} \)) using post-training quantization (PTQ). For FPGA-based soft sensors, the best full-precision (\( \text{PT}_{FP32} \)) model was obtained under identical experimental conditions but within PT framework. Notably, the quantized (\( \text{PT}_{quant} \)) model was generated using fixed-point (6,8) quantization-aware training (QAT), where 8 represents the total bit width and 4 is allocated for fractional bits.

\subsection{Results and Evaluation}

Figure \ref{fig:results} illustrates the MSE of our MLP models with various configurations of hidden neurons, trained under both TF and PT frameworks. The MSEs of our models are competitive compared to the study by Noori et al.~\cite{noori2020non}, which achieved a test MSE as low as 78.13 under a similar setup. Within the TF framework, both \( \text{TF}_{FP32} \) and \( \text{TF}_{quant} \) models exhibit minor fluctuations in MSE across different numbers of hidden neurons, indicating consistent performance. In contrast, within the PT framework, \( \text{PT}_{FP32} \) models show more substantial variance in MSE. Notably, except for the model with 10 hidden neurons, \( \text{PT}_{FP32} \) models perform better compared to their counterparts. However, a significant increase in MSE is observed among the \( \text{PT}_{quant} \) models, suggesting accuracy degradation due to the applied quantization strategy. Furthermore, both \( \text{PT}_{FP32} \) and \( \text{PT}_{quant} \) models exhibit considerable MSE variations with different hidden neuron counts, underscoring their sensitivity to model size adjustments.

\begin{figure}[!htb]
\centering
\begin{tikzpicture}
\begin{axis}[
    width=0.62\textwidth,
    height=0.40\textwidth,
    xlabel={Hidden Size},
    ylabel={MSE},
    xlabel={Neurons in the hidden layer},
    ylabel={MSE},
    xmin=0, xmax=120,
    ymin=50, ymax=75,
    xtick={10,30,60,120},
    ytick={50,55,60,65,70,75},
    ymajorgrids=true,
    grid style=dashed,
    legend pos=north east,
    legend style={at={(1.02,1)},anchor=north west
    },
]




\addplot[color=blue, line width=1pt,mark=*,mark options={fill=blue}]
    coordinates {(10,58.15)(30,56.97)(60,55.68)(120,54.21)};
\addlegendentry{ $\text{TF}_{FP32}$}

\addplot[color=blue, line width=1pt, dashed,mark=o,mark options={solid}]
    coordinates {(10,58.81)(30,56.33)(60,56.49)(120,53.69)};
\addlegendentry{$\text{TF}_{quant}$}

\addplot[color=red, line width=1pt,mark=*,mark options={fill=red}]
    coordinates {(10,58.51)(30,53.28)(60,51.83)(120,50.65)};
\addlegendentry{$\text{PT}_{FP32}$}

\addplot[color=red, line width=1pt, dashed,mark=o,mark options={solid}]
    coordinates {(10,73.96)(30,63.69)(60,64.14)(120,62.04)};
\addlegendentry{$\text{PT}_{quant}$}

\end{axis}
\end{tikzpicture}
\caption{MSE Values Under TF and PT Frameworks Across Various Hidden Neurons}
\label{fig:results}
\end{figure}
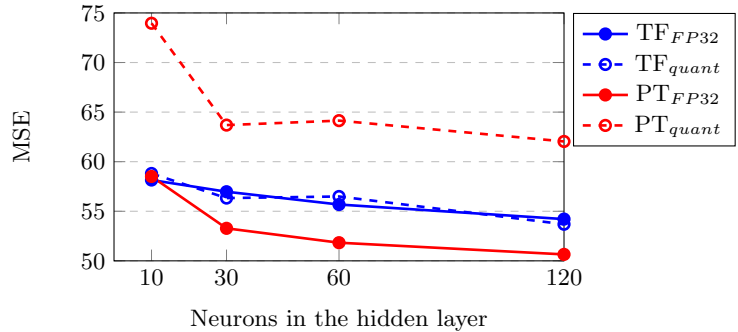
\begin{table}[!htb]
\centering
\setlength{\tabcolsep}{8pt}
\begin{tabular}{c|cccc}
\specialrule{.15em}{.1em}{.1em}
\textbf{Hardware} & \textbf{\begin{tabular}[c]{@{}c@{}}Hidden \\ Neurons \end{tabular}} & \textbf{\begin{tabular}[c]{@{}c@{}}Inference \\ time ($\mu$s)\end{tabular}} & \textbf{\begin{tabular}[c]{@{}c@{}}Power \\ (mW) \end{tabular}} & \textbf{\begin{tabular}[c]{@{}c@{}}Energy \\ ($\mu$J/inference)\end{tabular}} \\
\hline
\multirow{4}{*}{\textbf{MCU}} 
& {10} & 29.58 & 119 & 3.52 \\
& {30} & 56.35 & 119 & 6.71 \\
& {60} & 96.48 & 119 & 11.48 \\
& {120} & 176.89 & 119 & 21.05 \\
\hline
\multirow{4}{*}{\textbf{FPGA}} 
& {10} & 1.04 & 28 & 0.03 \\
& {30} & 3.04 & 29 & 0.09 \\
& {60} & 6.04 & 29 & 0.18 \\
& {120} & 12.04 & 29 & 0.35 \\
\specialrule{.15em}{.1em}{.1em}
\end{tabular}
\caption{Performance Comparison with Various Hidden Neurons}
\vspace{-10pt}
\label{tab:results}
\end{table}

These insights emphasize the crucial role of selecting the appropriate framework, quantization strategy, and model size in enhancing on-device soft sensors. Building on these findings, Table \ref{tab:results} presents the performance metrics of soft sensors deployed on MCUs, as well as the estimated performance of FPGA-based soft sensors derived from Vivado simulations. As the number of hidden neurons escalates, the inference time on MCUs increases. Specifically, with 60 neurons, the inference time approaches 96.48 $\mu$s, closely aligning with the real-time requirement of 100 $\mu$s at a 10 kHz sampling rate. In stark contrast, the inference on FPGA is expedient, being up to 28.44$\times$ faster than its MCU counterpart depending on the neuron count, thereby satisfying real-time requirements even with 120 hidden neurons. While the power consumption of the inference on MCU remains consistently at 119 mW regardless of neuron number, the consumption of the inference on FPGA peaks at a mere 29 mW. This combination of rapid inference speed and low power consumption endows the FPGA with superior energy efficiency per inference relative to MCU. 


\section{Conclusion and Future Work}
\label{sec:conclusion}

This study highlights the viability and efficiency of on-device soft sensors for real-time fluid flow estimation. Through rigorous experimentation with various MLP models on MCUs and FPGAs, we conclusively demonstrate the superior inference speed and energy efficiency afforded by FPGA deployments. Nonetheless, the $\text{PT}_{\text{quant}}$ models necessitate significant performance improvements when compared to their $\text{TF}_{\text{quant}}$ counterparts. Our findings also shed light on the subtle trade-offs and advantages of different hardware platforms and model configurations, providing valuable insights for future research and applications. Looking forward, we aim to refine and improve quantization strategies for FPGA implementations. This endeavor seeks to mitigate accuracy losses due to the quantization process, all while continuing our dedication to developing robust and dependable on-device AI applications designed for fluid flow estimation tasks.

\section*{Acknowledgments}
The authors gratefully acknowledge the financial support provided by the Federal Ministry for Economic Affairs and Climate Action of Germany for the RIWWER project (01MD22007C). 
\bibliographystyle{splncs04}
\bibliography{references}
\end{document}